\documentclass{article}

\usepackage[final]{corl_2026} 
\usepackage{booktabs}
\usepackage{amsmath}
\usepackage{amssymb}
\usepackage{enumitem}
\usepackage{graphicx}
\usepackage{pgfplots}
\usepackage{wrapfig}
\usepackage{needspace}
\usepackage{float}
\pgfplotsset{compat=1.18}

\title{FAWAM: Force-Aware World Action Models for Closed-Loop Contact-Rich Manipulation}

\author{
    Haotian He$^{*,1}$, Zeyu Yan$^{*,2}$, Qipeng Liu$^{2}$, Ning Guo$^{2}$, Wenzhao Lian$^{2,\dagger}$\\
    $^{1}$ Peking University \quad
    $^{2}$ Shanghai Jiao Tong University\\
    ${*}$ Equal Contribution;$^{\dagger}$ Corresponding author\\
    \href{https://fawam.github.io/}{Project Page: \texttt{https://fawam.github.io}}
}

\begin{document}
\maketitle

\begin{abstract}
    Force signals provide critical interaction cues for contact-rich robotic manipulation. However, existing methods mostly use force as an additional observation modality, without fully exploiting its role in modeling future interaction dynamics or guiding execution-time feedback correction. In this paper, we propose \textbf{FAWAM}, a force-aware world action model that incorporates force information at three levels: perception, prediction, and closed-loop execution. FAWAM first encodes historical 6-axis force/torque signals to modulate action generation, then jointly predicts future actions and end-effector wrenches to explicitly model contact evolution. It further introduces a residual correction module that uses the predicted wrench trajectory as an execution-time reference to refine actions online based on real-time force feedback. Real-world experiments across multiple contact-rich tasks show that FAWAM improves the average success rate by \textbf{36.25}\% over vision-only baselines and \textbf{21.25}\% over existing force-aware baselines, demonstrating the effectiveness of our force-aware framework for robust contact-rich manipulation.
\end{abstract}

\keywords{World Action Model, Force-Aware Policy, Contact-Rich Manipulation}

\section{Introduction}
Contact-rich manipulation tasks, such as wiping, peeling and pushing, are ubiquitous in human activities and real-world robotic applications. These tasks require the robot to continuously regulate its interaction with the environment.
However, relying solely on visual observations are often insufficient, since critical interaction states, such as weak contact, excessive pressure or jamming, are difficult to infer from images. Consequently, force and tactile feedback provide essential complementary signals for capturing contact states that are ambiguous from vision alone~\citep{lee2019making,zhao2025touch,zheng2026omnivta,yu2025forcevla,chen2025dexforce,li2026tacot,xu2026vt}.

Recent works~\citep{fang2026forcepolicy,he2024foar,zhang2025tavla,sun2024force} have incorporated force sensing into visuomotor models and vision-language-action (VLA) models, demonstrating improved performance on contact-rich tasks. However, these methods mainly use force as an additional observation modality for policy learning~\citep{yu2025forcevla,hou2024adaptive}, or introduce force feedback into low-level controllers for reactive control~\citep{xue2025rdp,li2026favla}. Although these approaches partially compensate for the incompleteness and latency of visual observations, they do not explicitly model the evolution of future contact states, such as contact forces.
As a result, they lack predictive wrench references to guide execution-time corrections.


In contrast, humans performing contact-rich tasks continuously anticipate the contact outcomes of their motions and adjust when sensed feedback deviates from expectation. This suggests that contact-rich manipulation is not purely a reactive observation-to-action mapping, but also a proactive prediction paradigm: the robot needs to anticipate how visual and contact states evolve under its planned action sequence. World models~\citep{liao2025genie,li2026lingbotva,pai2025mimic,bi2025motus,yuan2026fast}, which capture future state evolution, provide a natural framework for contact prediction and contact-aware action generation. Yet, prediction alone does not ensure robust execution. Predicted contact signals should further serve as online references during execution, allowing the robot to detect contact deviations and correct its actions before failures occur.



\begin{figure}[h]
    \centering
    \includegraphics[width=1.0\linewidth]{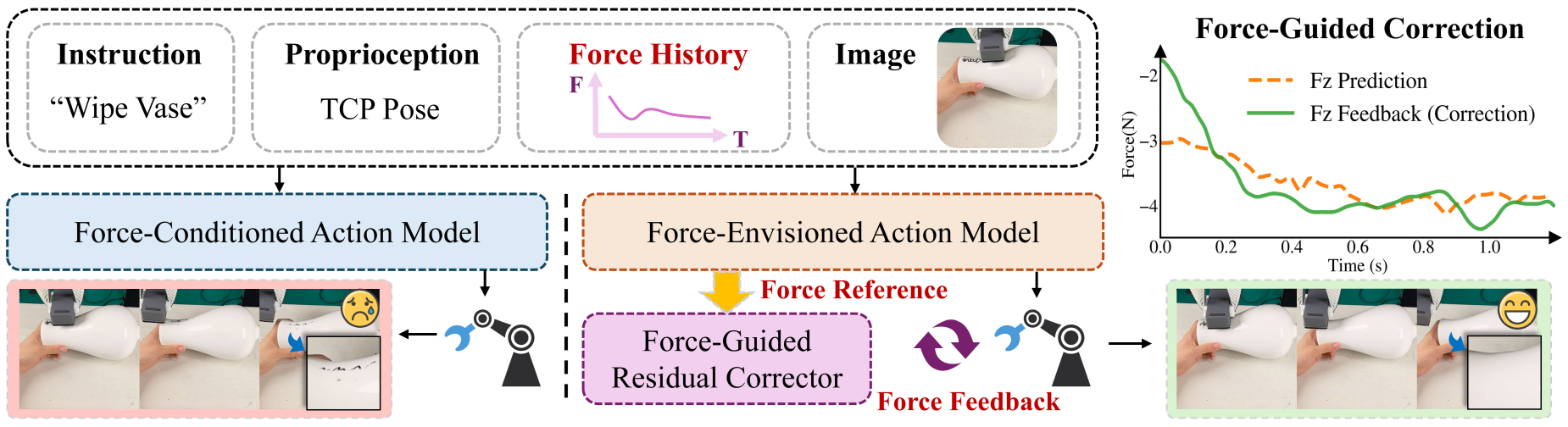}
    \caption{\textbf{Comparison between force-conditioned prediction and our method.} \textit{(left)} Adding force as additional observation alone fails to correct contact deviation while \textit{(right)} systematic force integration in three levels restores proper contact for successful wiping.}
    \label{fig:teaser}
\end{figure}

Built upon these observations, we propose \textbf{FAWAM}, a \textbf{F}orce-\textbf{A}ware \textbf{W}orld \textbf{A}ction \textbf{M}odel for contact-rich manipulation. As illustrated
in Figure~\ref{fig:teaser}, FAWAM systematically  leverages force signals at three levels: contact-state estimation, consequence prediction and execution-time correction. Specifically, we design a \textbf{Force-Envisioned Action Model} that encodes recent force/torque histories for local contact-state estimation and jointly predicts future actions and end-effector force signals to model contact evolution. Further, during action chunk execution, we introduce a \textbf{Force-Guided Residual Corrector} that uses the discrepancy between predicted and sensed force feedback to refine subsequent actions online. This systematic force integration yields robust behaviors adaptable to dynamic environmental variations, validated by empirical results across multiple contact-rich manipulation tasks.

Our contributions are summarized as follows:
\begin{itemize}[leftmargin=*, itemsep=0pt, topsep=1pt, parsep=0pt]
\item We propose FAWAM, a force-aware world action model that leverages force signals throughout perception, prediction and execution levels to enable robust contact-rich manipulation.
\item We introduce a force-guided residual correction mechanism that uses the discrepancy between predicted forces and sensed force feedback to correct actions during execution.
\item We validate our method on various contact-rich tasks, achieving \textbf{36.25}\% and \textbf{21.25}\% gains over the best vision-only and force-aware baselines. Extensive experiments demonstrate the superiority of our method and its robustness to external disturbances and diverse contact conditions.
\end{itemize}


\section{Related Work}
\textbf{Force-Aware Manipulation Policies.} 
Recent works~\citep{fang2026forcepolicy,sun2024force,pai2025mimic,buamanee2024bi,seo2025equicontact,zhou2025admittance} have explored various strategies to incorporate force signals into manipulation policies. FoAR~\citep{he2024foar} weights visual and force features with a future-contact predictor. 
ACP~\citep{hou2024adaptive} learns spatiotemporal approximate compliance from demonstrations to limit contact force while preserving tracking. 
FACTR~\citep{factr} uses force-attending curriculum training for contact-rich imitation learning. RDP~\citep{xue2025rdp} combines a low-frequency latent diffusion policy with a high-frequency fast policy for closed-loop force refinement. Force Policy~\citep{fang2026forcepolicy} decouples force regulation and motion execution through an interaction frame for hybrid force-position control. 
Force-enhanced VLAs extend pretrained policy backbones with contact signals: ForceVLA~\citep{yu2025forcevla} performs force fusion through a mixture-of-experts representation, while TA-VLA~\citep{zhang2025tavla} studies auxiliary torque prediction for physically grounded features learning. Although these methods incorporate force feedback, they typically do not explicitly model the evolution of future contact-related force signals, nor do they use such predictions to guide closed-loop action correction during execution.

\textbf{World Models for Action Generation.}
Benefiting from the ability to predict future dynamics, world action models have recently developed rapidly as a promising paradigm for robot policy learning~\citep{hafner2023dreamerv3,liao2025genie,wang2026wam,shen2026videovla,hou2026world,lyu2025dywa,World-VLA-Loop, chandra2025diwa, zheng2025flare,ye2026gigaworld}. Existing works follow three directions. Two-stage methods separate video prediction from action decoding: VPP~\citep{hu2024video} extracts predictive features from intermediate video diffusion states, while mimic-video~\citep{pai2025mimic} shows that video prediction quality closely influences policy quality. Joint-generation methods predict visual futures and actions together: WorldVLA~\citep{cen2025worldvla} shares parameters between world modeling and action prediction. Cosmos Policy~\citep{kim2026cosmospolicy} casts actions, proprioception, future states, and values as video-like tokens for planning. DreamZero~\citep{ye2026dreamzero} scales autoregressive video-action modeling for zero-shot and cross-embodiment generalization. Unified models~\citep{li2025unified,zhu2025unified} further merge action generation, video generation, forward and inverse dynamics together: 
LingBot-VA~\citep{li2026lingbotva} performs token-level video-action autoregression. Motus~\citep{bi2025motus} learns embodiment-agnostic motion latents from optical flow. Despite these advances, existing methods mainly model visual or latent state evolution, leaving force-related prediction and correction underexplored.

\textbf{On-Policy Imitation and Corrective Data Aggregation.}
Behavior cloning is vulnerable to covariate shift, as policies are trained on expert-induced states but deployed under their own induced distributions~\citep{seo2024mitigating}. DAgger~\citep{ross2011dagger} reduces this mismatch by aggregating expert labels on learner-visited states, and robotic variants adapt this idea to interactive imitation learning through human interventions and corrective demonstrations~\citep{kelly2018hgdagger,spencer2020interventions,mandlekar2020human,liu2022robot,hoque2021lazydagger,hoque2021thriftydagger,wu2025robocopilot}.
For contact-rich manipulation, interventions are more challenging because they must preserve ongoing contact and avoid abrupt force discontinuities. CR-DAgger~\citep{xu2025crdagger} addresses these issues by collecting smooth delta corrections with force feedback through a compliant intervention interface. 
Such corrective data naturally aligns with residual policy learning, which adds learned residual actions to a nominal controller to correct local execution errors~\citep{johannink2019residual,ankile2024residual,yuan2024policydecorator,he2025asap,haldar2023teach,guzey2024see}. This allows the base policy to preserve routine behavior while the residual module focuses on local execution errors. 
Building on this line, we focuses on integrating residual correction with force-aware policies for contact-rich manipulation.

\section{Method}
\label{sec:method}
\subsection{Overview}
\begin{figure}
    \centering
    \includegraphics[width=1.0\linewidth]{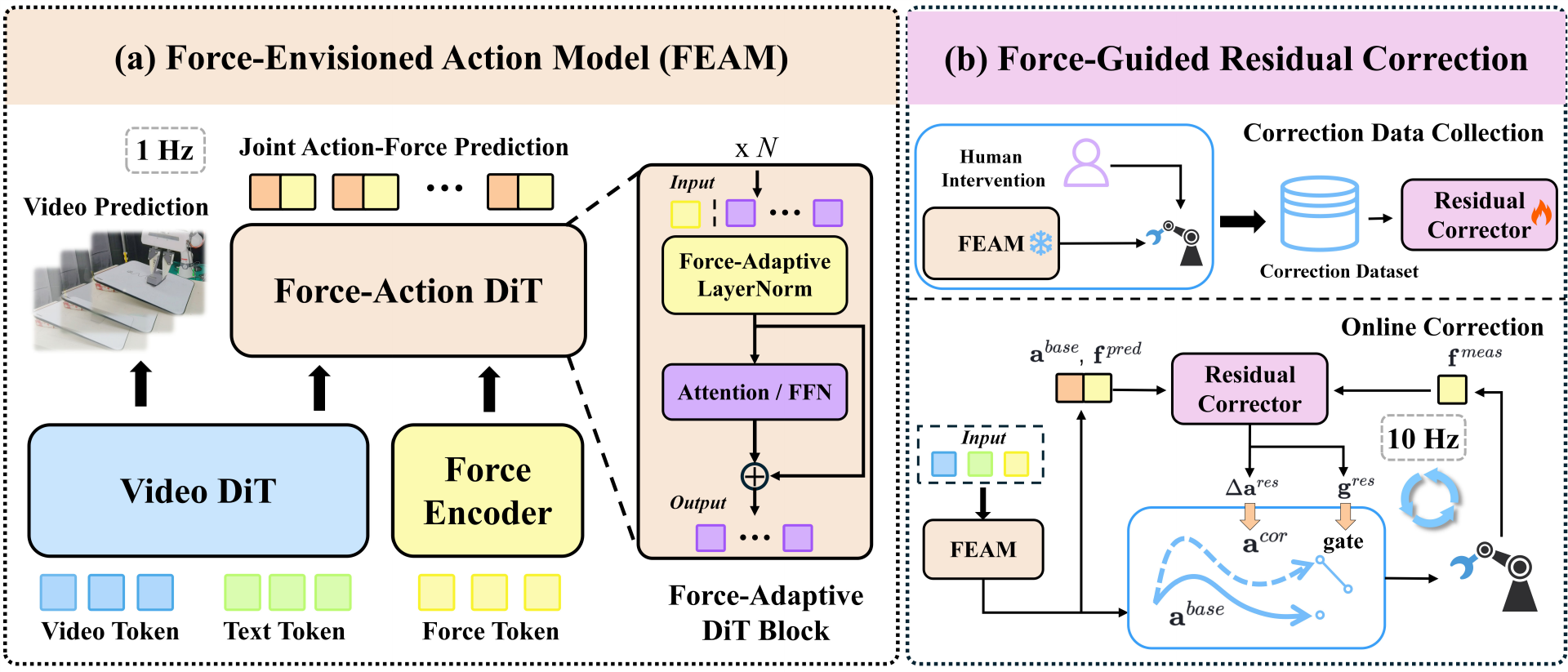}
    \caption{\textbf{Overview of FAWAM.} 
FAWAM incorporates force signals through force-conditioned action generation, future force prediction, and online residual correction. 
(a) The Force-Envisioned Action Model conditions on visual, language and force inputs to jointly predict action chunks and future force trajectories. 
(b) The Force-Guided Residual Correction module learns from human interventions and corrects the base action online with a residual action and an intervention gate.}
    \label{fig:FAWAM}
\end{figure}

Inspired by how humans sense, anticipate and react to contact during manipulation, we propose FAWAM, a closed-loop force-aware world action model for contact-rich tasks. As illustrated in Figure~\ref{fig:FAWAM}, FAWAM consists of a Force-Envisioned Action Model running at 1~Hz and a Force-Guided Residual Corrector running at 10~Hz. 
The Force-Envisioned Action Model encodes recent force/torque histories into compact contact features and injects them into action generation through AdaLN-Zero-style modulation. It also jointly predicts future action and force trajectories, encouraging the model to learn the coupling between robot motions and their contact consequences. During execution, the Force-Guided Residual Corrector uses the predicted force trajectory as guidance for online adaptation to unexpected contact changes within an action chunk. By comparing the predicted force reference with real-time force feedback, it provides timely residual corrections and improves execution-time robustness.

\subsection{Force-Envisioned Action Model}
To provide the world action model with local contact-state awareness, we encode recent high-frequency end-effector wrench histories into a compact contact feature. Let $\mathbf{w}_t \in \mathbb{R}^{6}$ denote the wrist force/torque measurement at time $t$, then the contact feature $\mathbf{c}_t$ can be defined as
\begin{equation}
    \mathbf{c}_t = \mathrm{Enc}_f(\mathbf{w}_{t-H_f+1:t}),
\end{equation}
where $\mathrm{Enc}_f$ is a lightweight MLP jointly trained with the action decoder.

Building upon GE-Act~\citep{liao2025genie}, which decodes action trajectories from future visual latent representations, we inject $\mathbf{c}_t$ into the action generation pathway through force-conditioned AdaLN-Zero modulation. Let $\mathbf{u}_t^{(i)}=
    (\boldsymbol{\beta}_{\mathrm{msa}}, \boldsymbol{\gamma}_{\mathrm{msa}}, \mathbf{g}_{\mathrm{msa}},
    \boldsymbol{\beta}_{\mathrm{mlp}}, \boldsymbol{\gamma}_{\mathrm{mlp}}, \mathbf{g}_{\mathrm{mlp}})$ denote the original AdaLN-Zero parameters in the $i$-th action DiT block, including the shift, scale, and gate values for both attention and MLP branches.
We predict a force-conditioned offset and add it to the original AdaLN-Zero parameters:
\begin{equation}
    \bar{\mathbf{u}}_t^{(i)}
    =
    \mathbf{u}_t^{(i)}
    +
    \mathrm{MLP}_{f}^{(i)}(\mathbf{c}_t).
\end{equation}
The modulated parameters $\bar{\mathbf{u}}_t^{(i)}$ are then used in the standard AdaLN-Zero computation of the action DiT block. The last layer of $\mathrm{MLP}_{f}^{(i)}$ is zero-initialized, so the model initially recovers the original action decoder and gradually learns force-adaptive refinements during training.

Beyond conditioning on historical wrenches, we further supervise the decoder to predict future wrenches. Instead of attaching an isolated force-prediction head, we replace the original action projection with a joint linear projection that outputs concatenated action and wrench trajectories. This encourages the decoder to learn a shared representation for action generation and contact prediction, capturing the coupling between robot motions and their physical contact consequences.

We adopt a two-stage training strategy. 
In the first stage, we fine-tune the video generation model
based on GE-Base~\citep{liao2025genie} with a latent-space flow-matching objective. Specifically,
given historical multi-view visual latents $\mathbf{I}_{t-K:t}$, language instruction $\mathbf{L}_t$, and future visual latent sequence $\mathbf{O}_t=[\mathbf{I}_{t+1},\cdots,\mathbf{I}_{t+H}]$, 
we construct the interpolated input $\hat{\mathbf{O}}_t^{\alpha}=\alpha\mathbf{O}_t+(1-\alpha)\boldsymbol{\epsilon}^{O}$, where $\alpha\in[0,1]$ and $\boldsymbol{\epsilon}^{O}\sim\mathcal{N}(\mathbf{0},\mathbf{I})$. 
The video generation model is optimized by
\begin{equation}
    \mathcal{L}_{stage1}=\mathbb{E}
    \left[
    \left\|
    v_{\psi}^{O}
    \left(
        \hat{\mathbf{O}}_t^{\alpha},
        \alpha,
        \mathbf{L}_t,
        \mathbf{I}_{t-K:t}
    \right)-
    \left(
        \mathbf{O}_t-\boldsymbol{\epsilon}^{O}
    \right)
    \right\|_2^2
    \right].
    \label{eq:video_fm_loss}
\end{equation}
In the second stage, we freeze the video world model and train the force-aware action decoder. 
Let $\tilde{\mathbf{o}}_t$ denotes the visual feature from the adapted video world model, $\mathbf{q}_t$ denote the proprioceptive state. 
We concatenate the future action trajectory $\mathbf{A}_t$ and future wrench trajectory $\mathbf{F}_t$ as $\mathbf{Z}_t=[\mathbf{A}_t,\mathbf{W}_t]$, and construct $\hat{\mathbf{Z}}_t^{\alpha}=\alpha\mathbf{Z}_t+(1-\alpha)\boldsymbol{\epsilon}^{Z}$ with $\boldsymbol{\epsilon}^{Z}=[\boldsymbol{\epsilon}^{A},\boldsymbol{\epsilon}^{W}]$. 
The decoder predicts the joint velocity $[\hat{\mathbf{v}}_t^{A},\hat{\mathbf{v}}_t^{W}]=v_{\theta}^{Z}(\hat{\mathbf{Z}}_t^{\alpha},\alpha,\tilde{\mathbf{o}}_t,\mathbf{q}_t,\mathbf{c}_t)$, then the action and force losses are defined as
\begin{equation}
    \mathcal{L}_{action}
    =
    \mathbb{E}
    \left[
    \left\|
    \hat{\mathbf{v}}_t^{A}
    -
    \left(
        \mathbf{A}_t-\boldsymbol{\epsilon}^{A}
    \right)
    \right\|_2^2
    \right],\quad
    \mathcal{L}_{force}
    =
    \mathbb{E}
    \left[
    \left\|
    \hat{\mathbf{v}}_t^{W}
    -
    \left(
        \mathbf{W}_t-\boldsymbol{\epsilon}^{W}
    \right)
    \right\|_2^2
    \right].
\end{equation}
The final action-force training objective is
$    \mathcal{L}_{stage2}
    =
    \mathcal{L}_{action}
    +
    \lambda_F \mathcal{L}_{force}$,
where $\lambda_F$ balances the action generation and force prediction.
This formulation enables the decoder to learn future motion generation together with wrench prediction under the same flow-matching framework.

\subsection{Force-Guided Residual Correction}
The action chunk generated by the action decoder is executed open-loop within its horizon, making it difficult to react to unexpected contact variations, such as insufficient pressure, excessive force or sudden resistance. Therefore, we introduce a lightweight force-guided residual corrector that runs at 10~Hz and adjusts the remaining actions within the current chunk.

\subsubsection{Residual Corrector Design}
\label{Res_design}
The residual model takes base action chunk, predicted wrench reference and real-time wrench feedback as inputs and outputs residual action corrections. Specifically, we use the same force encoder as the world action model to encode recent wrench histories into the contact feature $\mathbf{c}_t$, with the encoder frozen during residual training. To expose contact deviations explicitly, we compute the wrench prediction discrepancy
\begin{equation}
    \mathbf{e}_{t-H_e+1:t}^{w}
    =
    \mathbf{w}_{t-H_e+1:t}^{meas}
    -
    \mathbf{w}_{t-H_e+1:t}^{pred},
\end{equation}
where $\mathbf{w}^{meas}$ denotes the measured 6D force/torque feedback, $\mathbf{w}^{pred}$ denotes the predicted wrench trajectory and $H_e$ denotes the history window used to compute the wrench prediction discrepancy. We also provide the residual model with the upcoming base action chunk $\mathbf{a}_{t+1:t+H_p}^{base}$ and predicted future wrench chunk $\mathbf{w}_{t+1:t+H_p}^{pred}$, which serve as previews of the intended motion and contact evolution. The current proprioceptive state $\mathbf{q}_t$ is included to account for the robot configuration.

Since residual corrections should only be applied when necessary, we introduce a gating head that predicts a scalar residual gate $g_t^{res}$. At inference time, the gate is converted into a binary activation mask $\kappa_t\in\{0,1\}$, where $\kappa_t=1$ if $g_t^{res}>\tau$ and $\kappa_t=0$ otherwise. We manually set a conservative threshold $\tau=0.8$ to prevent the residual model from perturbing nominal executions. The gate is trained with both intervention and non-intervention segments, as described in Sec.~\ref{sec:residual_training}. The correction process is formulated as
\begin{align}
    \Delta \mathbf{a}_{t+1:t+H_r}^{res},\; g_t^{res}
    &=
    \mathrm{Res}_{\phi}
    \left(
        \mathbf{e}_{t-H_e+1:t}^{w},
        \mathbf{c}_t,
        \mathbf{a}_{t+1:t+H_p}^{base},
        \mathbf{w}_{t+1:t+H_p}^{pred},
        \mathbf{q}_t
    \right), \\
    \mathbf{a}_{t+1:t+H_r}^{cor}
    &=
    \mathbf{a}_{t+1:t+H_r}^{base}
    +
    \kappa_t
    \Delta \mathbf{a}_{t+1:t+H_r}^{res},
\end{align}
where $\mathbf{a}_{t+1:t+H_r}^{cor}$ denotes the corrected action sequence and $\mathrm{Res}_{\phi}$ is implemented as a three-layer MLP for real-time inference.

\subsubsection{Correction Dataset Collection and Training}
\label{sec:residual_training}
We collect correction data with FACTR~\citep{factr}, a force-feedback leader-follower teleoperation system. The follower arm remains under impedance control for safe and compliant contact, while the leader arm tracks the policy-generated command during autonomous execution rather than mirroring the follower feedback. This keeps the command stream continuous when switching to human intervention, reducing action drift and force-feedback vibration. When abnormal contact is observed, the operator corrects the remaining action chunk through the leader arm. Then, the difference between the corrected command and the base policy command is recorded as the residual target. 

The training set contains both intervention segments and non-intervention policy rollouts. For non-intervention samples, the residual target is set to zero and the gate label is set to $0$. For intervention samples, the gate label is set to $1$. Each training batch contains a balanced mix of intervention and non-intervention samples, encouraging the model to correct abnormal executions while preserving nominal base-policy behavior. The residual corrector is trained with
\begin{equation}
    \mathcal{L}_{res}
    =
    \mathbb{E}
    \left[
    \left\|
        \Delta \mathbf{a}^{res}
        -
        \Delta \mathbf{a}^{gt}
    \right\|_2^2
    +
    \lambda_{gate}
    \operatorname{BCE}
    \left(
        g_t^{res},
        y_t
    \right)
    \right],
\end{equation}
where $\Delta \mathbf{a}^{gt}$ is the ground-truth residual correction, $y_t\in\{0,1\}$ indicates whether the sample comes from human intervention, and $\lambda_{gate}$ balances the residual regression and gate classification losses.

\section{Experimental Results}
\label{sec:result}

\subsection{Experimental Setup}

\textbf{Hardware Platform.}
All real-world experiments are conducted on a Franka Research 3 (FR3), a 7-DoF robotic arm. An ATI Axia80-M8 6-axis force/torque sensor is mounted at the wrist to measure interaction wrenches. Visual observations are captured by three Intel RealSense D435i cameras, including one wrist-mounted camera and two static third-person cameras from the front and top views.
For demonstration collection and human intervention, we build a force-feedback leader-follower teleoperation system based on FACTR~\citep{factr}. Force feedback from the follower arm is mapped to the leader arm as joint torques, allowing the operator to perceive contact forces during teleoperation. A system overview is provided in Appendix~\ref{app:hardware}.

\begin{figure}
    \centering
    \includegraphics[width=1.0\linewidth]{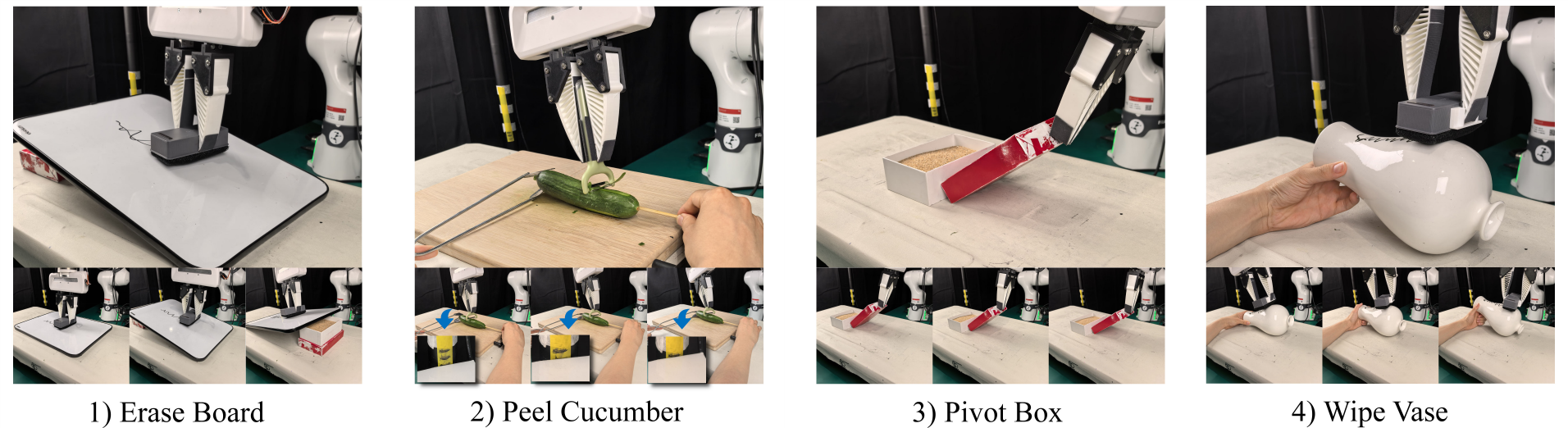}
    \caption{\textbf{Contact-rich tasks.} Each task is evaluated under diverse contact conditions induced by controlled changes in surface inclination, table height, or the position of the sand-filled box.}
    \label{fig:task}
\end{figure}
\textbf{Tasks and Data Collection.}
We evaluate our method on four contact-rich manipulation tasks: \textit{Erase Whiteboard}, \textit{Peel Cucumber}, \textit{Pivot Box} and \textit{Wipe Vase}, as shown in Figure~\ref{fig:task}. These tasks require sustained physical contact and force-sensitive execution under varying contact conditions. To improve data diversity, we introduce controlled variations for each task: changing the inclination angles of the whiteboard and vase, varying the table height for cucumber peeling, and changing the position of the sand-filled box for box pivoting. For each task, the base action model and all baseline policies are trained on the same 90 demonstrations. For each policy with residual correction, including FAWAM, we collect 20 additional human-intervention episodes per task under the same collection protocol, with on-policy corrections collected separately for each base policy. Multi-view camera observations and proprioceptive states are recorded at 30 Hz, while 6-axis force/torque measurements are recorded at 120 Hz.

\textbf{Baselines and Evaluation.}
We compare FAWAM with both vision-only and force-aware policy baselines.  For vision-only baselines, we evaluate $\pi_{0.5}$~\citep{black2025pi05} and GE-Act~\citep{liao2025genie}. $\pi_{0.5}$ is a generalist vision-language-action model without force input, while GE-Act is the vision-only backbone of FAWAM, using the same action prediction pipeline but without any force-related components. 
For force-aware baselines, we compare against ForceVLA~\citep{yu2025forcevla} and TA-VLA~\citep{zhang2025tavla}. ForceVLA incorporates historical force signals as an additional observation modality, while TA-VLA further uses force prediction as an auxiliary objective.
Under matched intervention budgets, we compare GE-Act + Res. (without wrench input) and TA-VLA + Res.
For each method, we perform 20 evaluation trials per task, with a maximum duration of 60 seconds for each trial. Detailed training settings are provided in Appendix~\ref{app:hardware}.

\subsection{Main Experiment Results}

As shown in Table~\ref{tab:evaluation_results}, FAWAM achieves the highest average success rate of 85\% among both vision-only and force-aware baselines.
Among policies without residual correction, vision-only baselines achieve average success rates below 50\% across the four tasks, while force-aware baselines improve performance to 63.75\% by incorporating force measurements or auxiliary force prediction.
Notably, FEAM, our base policy without residual correction, achieves an average success rate of 73.75\% without additional correction data, showing that the proposed force-envisioned action model already provides a strong base policy.
Residual correction raises FEAM's average success rate from 73.75\% to 85.00\%. Under matched intervention budgets, FAWAM also outperforms GE-Act + Res. and TA-VLA + Res. by 31.25\% and 8.75\%, respectively. This comparison confirms that FAWAM's gains are not solely due to access to intervention supervision. Execution-time comparisons and task-wise efficiency analysis are provided in Appendix~\ref{app:ablation_time}.


\begin{table}[t]
    \centering
    \caption{\textbf{Real-world evaluation results.}
Avg. SR denotes average success rate. Top block: policies without residual correction or additional intervention supervision.
Bottom block: policies with residual correction trained with matched intervention budgets.
All base policies are trained on the same demonstration dataset.
FEAM denotes our base policy without residual correction.}
    \label{tab:evaluation_results}
    \small
    \begin{tabular}{cccccc}
        \toprule
        Policy & Erase Board & Peel Cucumber & Pivot Box & Wipe Vase & Avg. SR \\
        \midrule
        $\pi_{0.5}$~\citep{black2025pi05} & 9/20 & 8/20 & 10/20 & 11/20 & 47.50\% \\
        GE-Act~\citep{liao2025genie} & 9/20 & 10/20 & 9/20 & 11/20& 48.75\% \\
        ForceVLA~\citep{yu2025forcevla} & 11/20 & 14/20 & 13/20 & 13/20 & 63.75\%\\
        TA-VLA~\citep{zhang2025tavla} & 10/20 & 17/20 & 14/20 & 10/20& 63.75\% \\
        FEAM & 14/20 & 15/20 & 16/20 & 14/20 & 73.75\% \\
        \midrule
        GE-Act + Res. & 11/20 & 10/20 & 10/20 & 12/20 & 53.75\% \\
        TA-VLA + Res. & 13/20 & \textbf{19/20} & 15/20 & 14/20 & 76.25\% \\
        FAWAM (ours) & \textbf{15/20} & \textbf{19/20} & \textbf{17/20} & \textbf{17/20} &\textbf{85.00\%} \\
        \bottomrule
    \end{tabular}
\end{table}

\subsection{Ablation Study}
\begin{wraptable}{r}{0.63\linewidth}
    \centering
    \caption{\textbf{Ablation study of force integration design.} \textit{Obs}: force-conditioned observation. \textit{Pre}: force-action joint prediction. \textit{Res}: residual correction.}
    \label{tab:ablation_study}
    \scriptsize
    \setlength{\tabcolsep}{3pt}
    \begin{tabular}{cccccccc}
        \toprule
        \textit{Obs} & \textit{Pre} & \textit{Res} & Erase Board & Peel Cucumber & Pivot Box & Wipe Vase & Avg. SR \\
        \midrule
        & & & 9/20 & 10/20 & 9/20 & 11/20 & 48.75\% \\
        $\checkmark$ & & & 12/20 & 12/20 & 9/20 & 12/20 & 56.25\% \\
        $\checkmark$ & $\checkmark$ & & 14/20 & 15/20 & 16/20 & 14/20 & 73.75\% \\
        & & $\checkmark$ & 12/20 & 15/20 & 12/20 & 15/20 & 67.50\% \\
        $\checkmark$ & $\checkmark$ & $\checkmark$ & 15/20 & 19/20 & 17/20 & 17/20 & \textbf{85.00\%} \\
        \bottomrule
    \end{tabular}
\end{wraptable}
\textbf{Force Integration Design.}
To analyze the role of different force-related designs in FAWAM, we conduct ablation studies in Table~\ref{tab:ablation_study}. Here, \textit{Obs} denotes adding wrench histories to the observation, \textit{Pre} denotes the joint action-force prediction and \textit{Res} denotes the residual correction module. For the \textit{Res}-only variant, no predicted wrench reference is available; therefore, the residual corrector only uses sensed force feedback as an additional input. We provide implementation details in Appendix~\ref{app:res_only}. Simply adding force signals to the observation yields only a 7.5\% improvement, suggesting that passive force conditioning is insufficient. Building on force-conditioned observations, joint action-force prediction brings a further 17.5\% improvement, indicating that future wrench prediction encourages physically grounded action representations. Residual correction alone also improves performance by 18.75\%, but remains below the full model without force-aware representation learning and predicted wrench guidance. Overall, these results show that joint prediction and residual correction provide complementary benefits beyond integrating force observation alone.

\begin{wraptable}{l}{0.42\linewidth}
    \centering
    \vspace{-1.0em}
    \caption{\textbf{Ablation study of the residual corrector design.}}
    \label{tab:residual_design}
    \scriptsize
    \setlength{\tabcolsep}{3pt}
    \begin{tabular}{lcc}
        \toprule
        Policy & Peel Cucumber & Wipe Vase \\
        \midrule
        FEAM + Res. (w/o wrench) & 15/20 & 15/20 \\
        FEAM + Res. (w/o gate) & 16/20 & 16/20 \\
        FAWAM & \textbf{19/20} & \textbf{17/20} \\
        \bottomrule
    \end{tabular}
\end{wraptable}
\textbf{Residual Corrector Design.}
Table~\ref{tab:residual_design} shows that removing wrench inputs reduces success rates on both tasks, supporting the importance of force feedback for correction. Disabling the gate ($\tau=0$) also degrades performance, suggesting that unnecessary residual activation may perturb nominal execution. A more detailed analysis of the gate threshold is provided in Appendix~\ref{app:residual_threshold}.

\subsection{Robustness under Execution-Time Perturbations}
\begin{figure}
    \centering
    \includegraphics[width=1.0\linewidth]{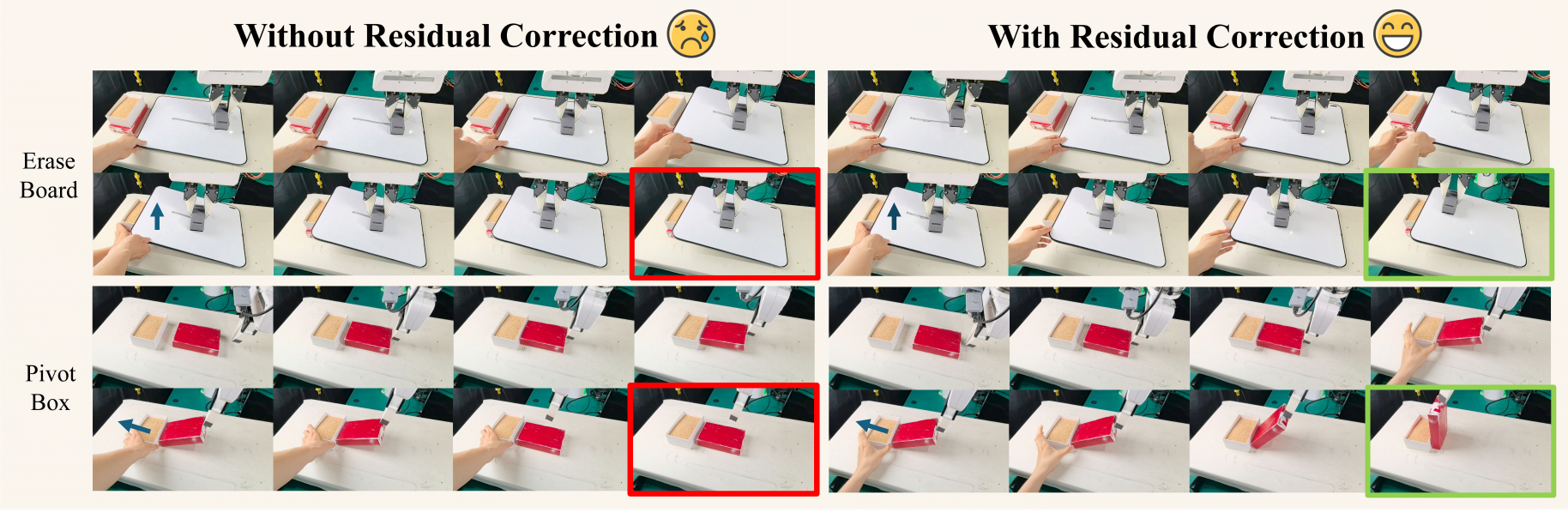}
    \caption{\textbf{Perturbation rollouts comparison.} With perturbations, force-guided residual correction help FAWAM recover stable contact, while the model without correction fails to adapt.}
    \label{fig:perturbation}
\end{figure}

\begin{wraptable}{r}{0.4\linewidth}
    \vspace{-1.5em}
    \centering
    \caption{\textbf{Results under perturbations.}}
    \label{tab:perturbations}
    \scriptsize
    \setlength{\tabcolsep}{2pt}
    \begin{tabular}{@{}ccccc@{}}
        \toprule
        Policy & \shortstack{Erase\\board} & \shortstack{Peel\\cucumber} & \shortstack{Slice\\cucumber} & \shortstack{Wipe\\vase} \\
        \midrule
        w/o. Correction model & 0/5 & 2/5 & 0/5 & 1/5 \\
        w/. Correction model & 3/5 & 5/5 & 4/5 & 4/5 \\
        \bottomrule
    \end{tabular}
\end{wraptable}
To further evaluate the role of the force-guided correction model, we test whether the policy can recover from unexpected contact changes during execution. 
Specifically, we manually introduce perturbations in four tasks: 
tilting the board during erasing, changing the cucumber 
height during peeling, moving the sand-filled box during pivoting, and changing the tail height of the vase during wiping. 
For each perturbation setting, we conduct five real-world trials. 
As shown in the Table~\ref{tab:perturbations}, adding the correction model consistently improves robustness under perturbations. 
Figure~\ref{fig:perturbation} shows representative perturbation rollouts. 
Without the correction model, the policy executes action chunks largely open-loop and fails to adapt to contact changes, causing jamming in board erasing and toppling in box pivoting. In contrast, the correction model uses discrepancies between predicted and sensed wrench to adjust subsequent actions, recover stable contact, and complete the task. Additional perturbation examples are provided in Appendix~\ref{app:perturbation_examples}.
These results show that force prediction alone is insufficient under execution-time disturbances, while the correction model closes the loop by turning force discrepancies into corrective behavior.

\section{Conclusion}
\label{sec:conclusion}
In this paper, we present FAWAM, a force-aware world action model for robust contact-rich manipulation. Instead of treating force as merely an additional observation modality, FAWAM uses force as both a predictive signal for contact-aware action generation and a corrective signal for execution-time feedback. We instantiate this idea with a Force-Envisioned Action Model and a Force-Guided Residual Corrector. Real-world experiments show that FAWAM achieves an 85.0\% average success rate, improving over the strongest vision-only and force-aware baselines by 36.25\% and 21.25\%, respectively. 
Ablation studies show that action-force prediction and residual correction bring complementary improvements, while perturbation experiments show that force-guided residual updates help recover from unexpected contact changes. These results highlight the importance of using force in both prediction and online correction for robust contact-rich manipulation.

\section{Limitation}
\label{sec:limitation}
Despite the improved robustness, our method still has several limitations. First, due to the lack of large-scale video datasets with synchronized force/torque signals, our current framework does not integrate force into the video generation model, but instead couples actions and wrenches in the action branch. Second, the residual corrector is trained from a limited number of human intervention episodes, which may not cover diverse failure modes and limits recovery under large deviations. More diverse correction data under richer perturbations and failure scenarios may improve the correction performance. 
Finally, end-effector wrench prediction provides only indirect contact information. 
We plan to address this in future work by integrating explicit contact-state representations, such as local contact frames and dense tactile signals, to provide more direct and generalized action guidance.

\clearpage
\acknowledgments{We thank Zehui Zhao and Yuxuan Zhao for their constructive feedback during discussions on model design. We also thank the reviewers for their valuable suggestions, which helped strengthen our experimental evaluation and comparative analysis.}


\bibliography{example}  

\clearpage
\appendix

\section{Appendix}

\subsection{Hardware and Training Details}
\label{app:hardware}
\paragraph{Hardware.}
Figure~\ref{fig:experimental_setup} shows the physical layout of the follower robot, FACTR leader robot, and RealSense cameras. The RealSense RGB streams are recorded at $640\times480$. During training, each frame is resized to $256\times192$ before being fed to the video world model. The ATI wrist force/torque sensor records 6-axis wrenches at 120~Hz. We apply a second-order Butterworth low-pass filter with a 10~Hz cutoff to suppress high-frequency sensor noise. For each policy query, the force encoder uses the most recent 0.5~s of wrench history, corresponding to 60 force/torque frames.

\paragraph{Training.}
Training is performed on 8 NVIDIA A100 GPUs. For video generation model fine-tuning and force-aware action decoder training stages, the total batch size is 64 and gradient accumulation is 1. Both stages use bfloat16 mixed precision, DeepSpeed ZeRO-2, gradient clipping with norm 1.0, and a constant-with-warmup learning-rate schedule with 1,000 warmup steps. We use AdamW with $\beta_1=0.9$, $\beta_2=0.95$, $\epsilon=10^{-8}$, and weight decay $10^{-5}$. The video world model is fine-tuned for 30,000 steps with the learning rate $3\times10^{-5}$. We then freeze the video world model and train the force-aware action decoder for 30,000 steps with the learning rate $1.5\times10^{-4}$. We use $\lambda_F=1$ to balance the action and future-wrench prediction losses. The force encoder is implemented as a three-layer MLP with a hidden width of 256 and is trained together with the force-aware action decoder.

The residual corrector is trained separately as a dual-head MLP with the frozen force encoder trained in the Force-Envisioned action model. Its inputs include the 60-frame filtered wrench history, predicted future wrench chunk, upcoming base action chunk, past wrench-tracking error, and current proprioceptive state. The MLP uses two shared hidden layers of width 256 with GELU activations and dropout 0.1, followed by a residual-action head and a gate head. We train it with a balanced batch of 128 samples containing 64 intervention and 64 non-intervention samples for 500 epochs. We train the residual corrector using AdamW with a learning rate of  $3\times10^{-4}$, $\beta_1=0.9$, $\beta_2=0.999$, $\epsilon=10^{-8}$, and a weight decay of $10^{-6}$. The gate loss weight is manually set to $\lambda_{\mathrm{gate}}=1$.

\IfFileExists{fig/experimental_setup.png}{
\begin{figure}[h]
    \centering
    \includegraphics[width=0.6\linewidth]{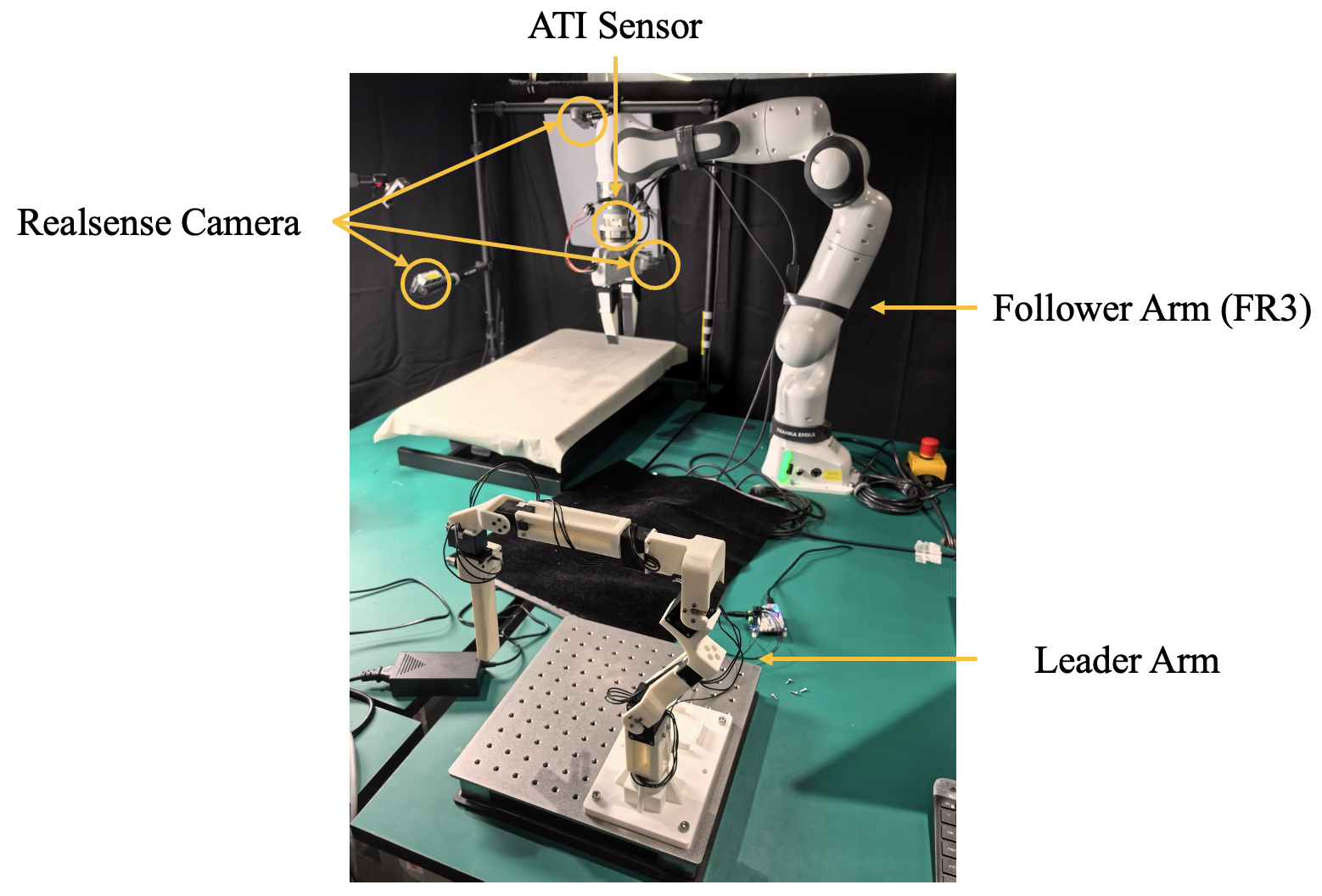}
    \caption{\textbf{Experimental environment.} The setup consists of a Franka robot for task execution, a FACTR leader robot for force-feedback teleoperation, and multiple Intel RealSense cameras for multi-view visual observation.}
    \label{fig:experimental_setup}
\end{figure}
}{}

\subsection{Residual-Only Ablation Details}
\label{app:res_only}
The \textit{Res}-only ablation isolates the effect of residual correction without the predicted wrench guidance used by the full model in Sec.~\ref{Res_design}. Since this variant does not include joint action-force prediction, it has no predicted wrench trajectory and therefore cannot construct either the future wrench preview or the measured-predicted wrench prediction discrepancy. We train a separate residual-only corrector that takes only the contact feature $\mathbf{c}_t$, the upcoming base action chunk, and the current proprioceptive state as inputs:
\begin{align}
    \Delta \mathbf{a}_{t+1:t+H_r}^{\mathrm{ro}},\; g_t^{\mathrm{ro}}
    &=
    \mathrm{Res}_{\phi}^{\mathrm{ro}}
    \left(
        \mathbf{c}_t,
        \mathbf{a}_{t+1:t+H_p}^{base},
        \mathbf{q}_t
    \right), \\
    \mathbf{a}_{t+1:t+H_r}^{cor}
    &=
    \mathbf{a}_{t+1:t+H_r}^{base}
    +
    \kappa_t^{\mathrm{ro}}
    \Delta \mathbf{a}_{t+1:t+H_r}^{\mathrm{ro}},
\end{align}
where the superscript $\mathrm{ro}$ denotes the residual-only ablation, $\mathbf{c}_t$ is encoded from the recent measured wrench history using the same frozen force encoder, and $\kappa_t^{\mathrm{ro}}$ is obtained by thresholding $g_t^{\mathrm{ro}}$ with the same gate threshold as the full residual corrector. The correction target, residual gate labels, and intervention-based training protocol remain the same as those in Sec.~\ref{sec:residual_training}.

\subsection{Execution-Time Analysis}
\label{app:ablation_time}

\begin{figure}[htbp]
    \centering
    \definecolor{basebar}{RGB}{216,203,198}
    \definecolor{obsbar}{RGB}{226,205,205}
    \definecolor{objobsbar}{RGB}{244,236,214}
    \definecolor{resbar}{RGB}{236,222,205}
    \definecolor{fullbar}{RGB}{232,137,119}
    \begin{tikzpicture}
        \begin{axis}[
            ybar=3pt,
            bar width=7pt,
            width=0.9\textwidth,
            height=0.25\textwidth,
            ymin=0,
            ymax=70,
            ylabel={Time (s)},
            ytick={0,20,40,60},
            symbolic x coords={Erase board,Peel cucumber,Pivot box,Wipe vase},
            xtick=data,
            xticklabel style={font=\scriptsize, align=center},
            yticklabel style={font=\scriptsize},
            ylabel style={font=\small},
            enlarge x limits=0.2,
            legend style={
                font=\scriptsize,
                draw=none,
                fill=white,
                fill opacity=0.85,
                text opacity=1,
                at={(0.98,0.98)},
                anchor=north east,
                legend columns=5,
                /tikz/every even column/.append style={column sep=2pt}
            },
            legend image code/.code={
                \path[#1, draw=none] (0cm,-0.08cm) rectangle (0.22cm,0.08cm);
            },
            ymajorgrids=true,
            grid style={dashed, gray!25},
            axis line style={gray!60},
            tick style={gray!60},
            point meta=explicit symbolic,
            nodes near coords={\pgfplotspointmeta},
            nodes near coords style={font=\fontsize{5}{5}\selectfont, anchor=south, yshift=-2pt, text=black},
            every node near coord/.append style={/pgf/number format/fixed, /pgf/number format/precision=1},
        ]
        \addplot+[fill=basebar, draw=none] coordinates {(Erase board,47.9) [47.9] (Peel cucumber,37.9) [37.9] (Pivot box,38.7) [38.7] (Wipe vase,42.6) [42.6]};
        \addplot+[
            fill=obsbar,
            draw=none,
            nodes near coords={
                \ifnum\coordindex=3
                    \raisebox{3pt}[0pt][0pt]{\pgfplotspointmeta}
                \else
                    \pgfplotspointmeta
                \fi
            }
        ] coordinates {(Erase board,35.2) [35.2] (Peel cucumber,59.0) [59.0] (Pivot box,40.0) [40.0] (Wipe vase,38.5) [38.5]};
        \addplot+[
            fill=objobsbar,
            draw=none,
            nodes near coords={
                \ifnum\coordindex=0
                    \raisebox{1pt}[0pt][0pt]{\pgfplotspointmeta}
                \else\ifnum\coordindex=3
                    \raisebox{1pt}[0pt][0pt]{\pgfplotspointmeta}
                \else
                    \pgfplotspointmeta
                \fi\fi
            }
        ] coordinates {(Erase board,38.2) [38.2] (Peel cucumber,45.4) [45.4] (Pivot box,42.6) [42.6] (Wipe vase,37.2) [37.2]};
        \addplot+[
            fill=resbar,
            draw=none,
            nodes near coords={
                \ifnum\coordindex=0
                    \raisebox{2pt}[0pt][0pt]{\pgfplotspointmeta}
                \else
                    \pgfplotspointmeta
                \fi
            }
        ] coordinates {(Erase board,39.6) [39.6] (Peel cucumber,56.3) [56.3] (Pivot box,37.8) [37.8] (Wipe vase,34.7) [34.7]};
        \addplot+[fill=fullbar, draw=none] coordinates {(Erase board,29.3) [29.3] (Peel cucumber,36.3) [36.3] (Pivot box,23.2) [23.2] (Wipe vase,31.8) [31.8]};
        \legend{Base, Obs, Obs+Obj, Res, Full}
        \end{axis}
    \end{tikzpicture}
    \caption{\textbf{Execution time comparison.} Lower values indicate faster completion with less inefficient contact adjustment.}
    \label{fig:ablation_time}
\end{figure}

Figure~\ref{fig:ablation_time} complements the success-rate results in Table~\ref{tab:ablation_study} by showing how different force-related components affect execution efficiency. The timing results should not be interpreted in isolation, since a shorter rollout can also arise from brittle open-loop behavior or early task termination. Instead, they are most informative when considered together with task success. Across the four tasks, the full FAWAM model achieves the best combination of high success rate and low completion time, indicating that the proposed force-aware prediction and residual correction do not merely improve robustness, but also reduce inefficient recovery behaviors during contact-rich execution.

The task-wise trends reveal how the benefit arises. On \textit{Erase Board}, maintaining stable normal contact is critical; variants without force-guided correction often lose contact or apply inappropriate pressure, causing repeated re-contact attempts. The full model completes the task faster by using force histories and residual updates to maintain consistent wiping pressure. On \textit{Peel Cucumber}, FAWAM combines the shortest average execution time with the highest success rate, suggesting that local force-guided adjustments improve task efficiency as well as reliability. On \textit{Pivot Box}, execution time is sensitive to whether the push direction and contact force produce effective object rotation. Joint action-force prediction helps generate physically consistent pushes, while residual correction prevents small force errors from accumulating into ineffective contacts. On \textit{Wipe Vase}, the curved surface requires continuous adaptation of contact pressure and tangential motion; the full model benefits from both sensed force histories and predicted wrench references to avoid prolonged slipping or over-pressing. Overall, the execution-time gains come from closing the loop around contact dynamics rather than simply producing faster nominal trajectories.


\subsection{Additional Perturbation Examples}
\label{app:perturbation_examples}
\begin{figure}[H]
    \centering
    \includegraphics[width=0.92\linewidth]{\detokenize{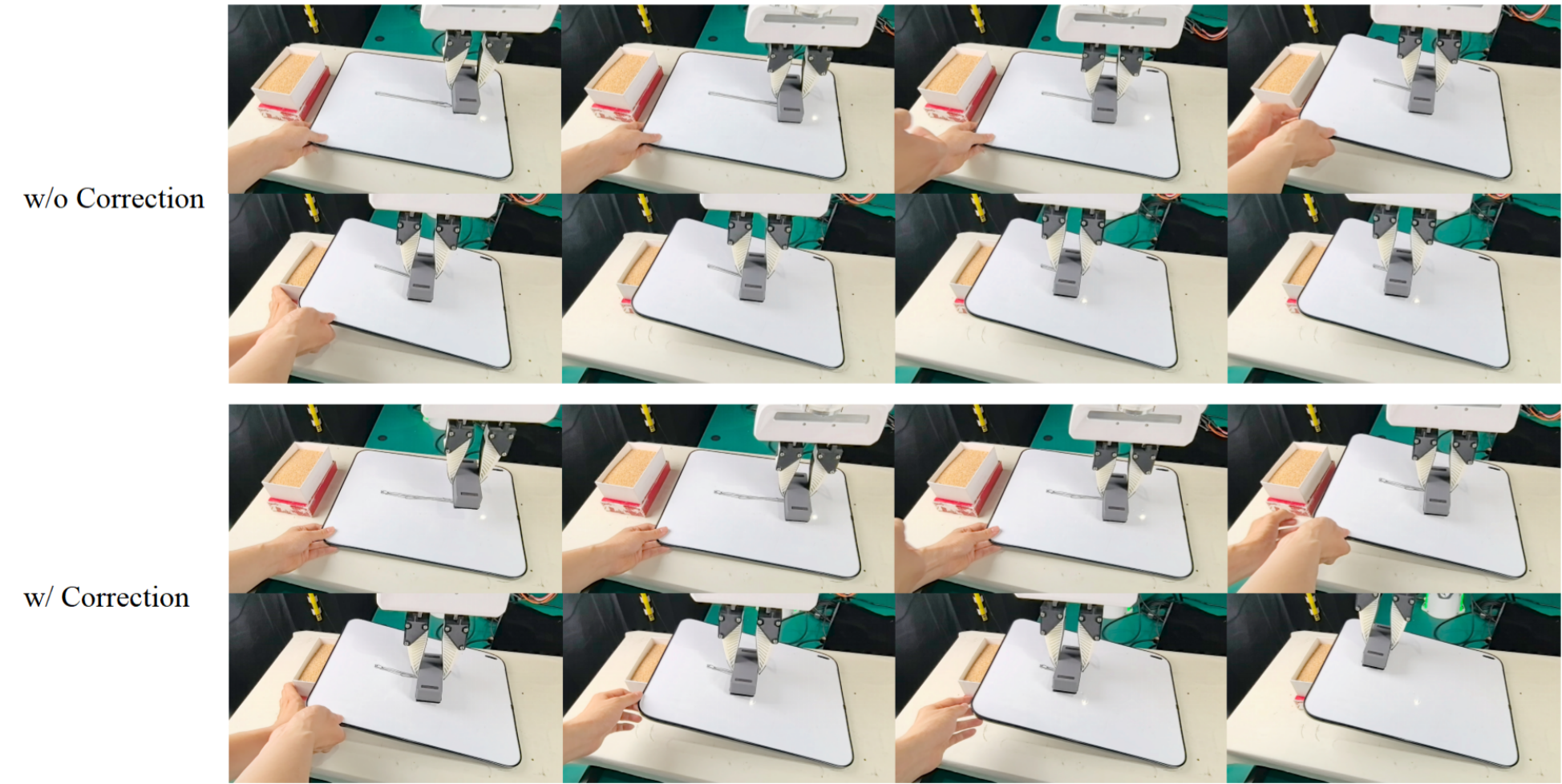}}
    \caption{\textbf{Comparative Correction Experiments for Erase Board.}}
    \label{fig:appendix_video_snapshot_1}
\end{figure}

\begin{figure}[H]
    \centering
    \includegraphics[width=0.92\linewidth]{\detokenize{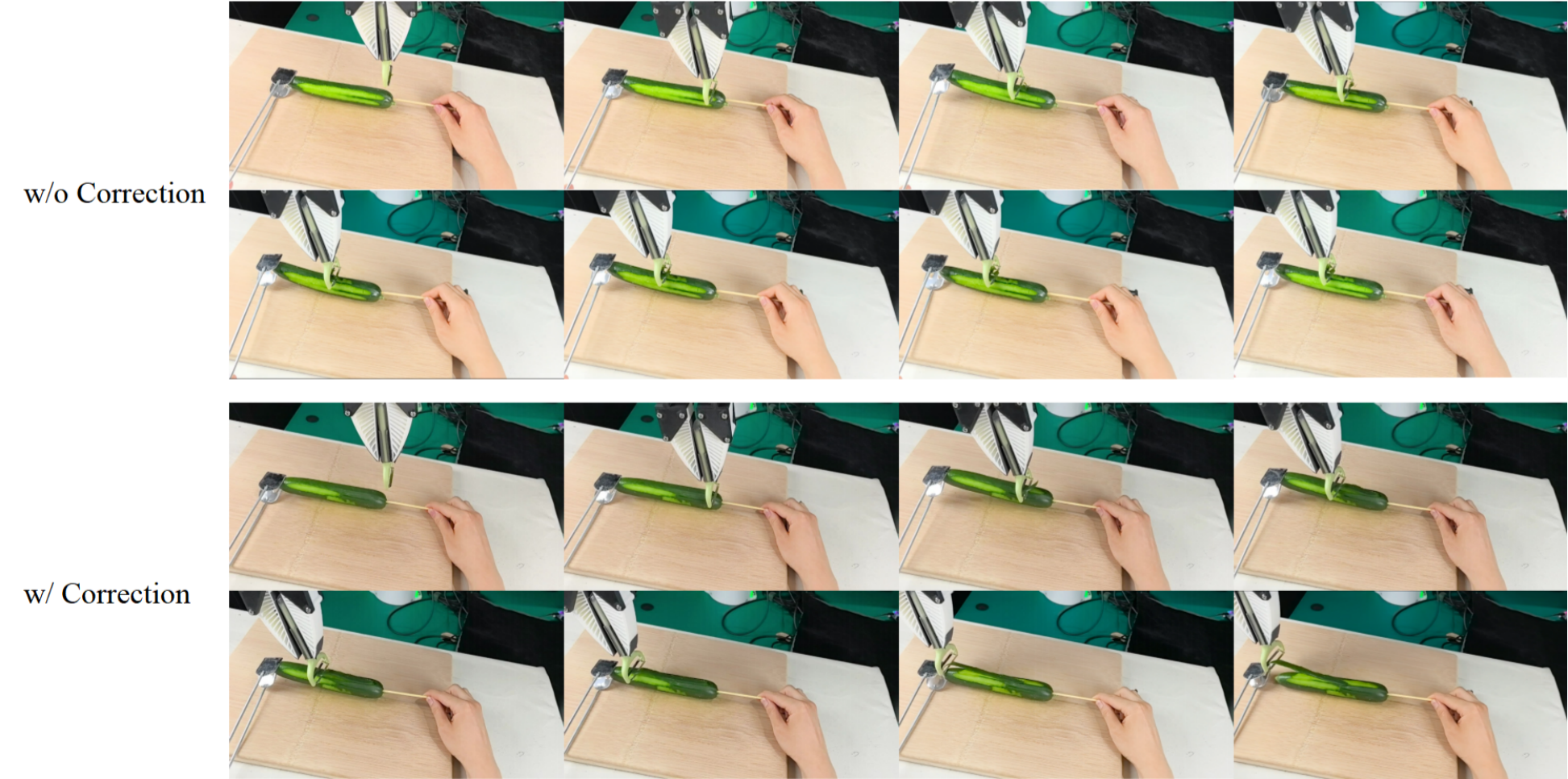}}
    \caption{\textbf{Comparative Correction Experiments for Peel Cucumber.} }
    \label{fig:appendix_video_snapshot_2}
\end{figure}

\begin{figure}[H]
    \centering
    \includegraphics[width=0.92\linewidth]{\detokenize{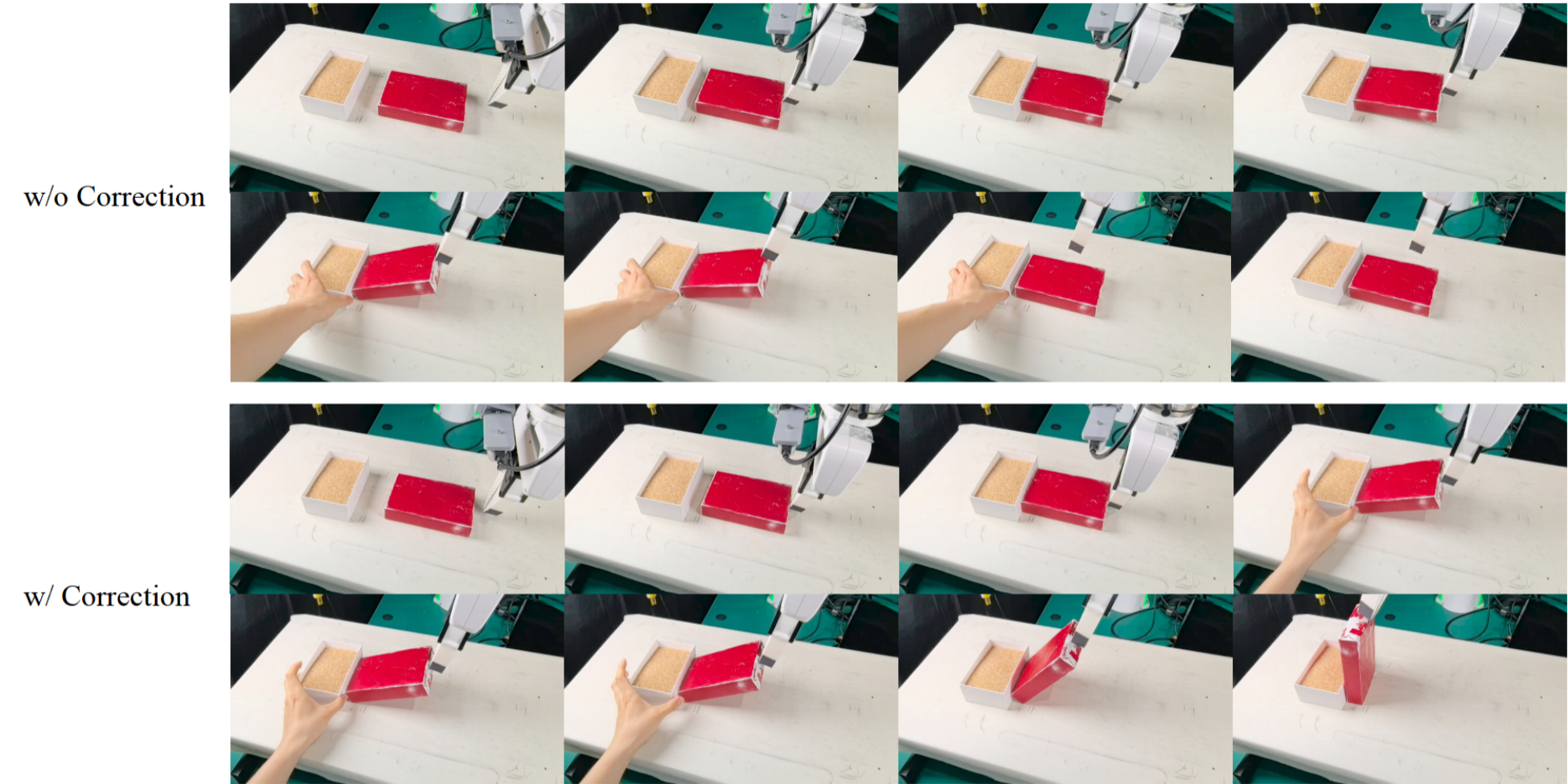}}
    \caption{\textbf{Comparative Correction Experiments for Pivot Box.} }
    \label{fig:appendix_video_snapshot_3}
\end{figure}

\begin{figure}[H]
    \centering
    \includegraphics[width=0.92\linewidth]{\detokenize{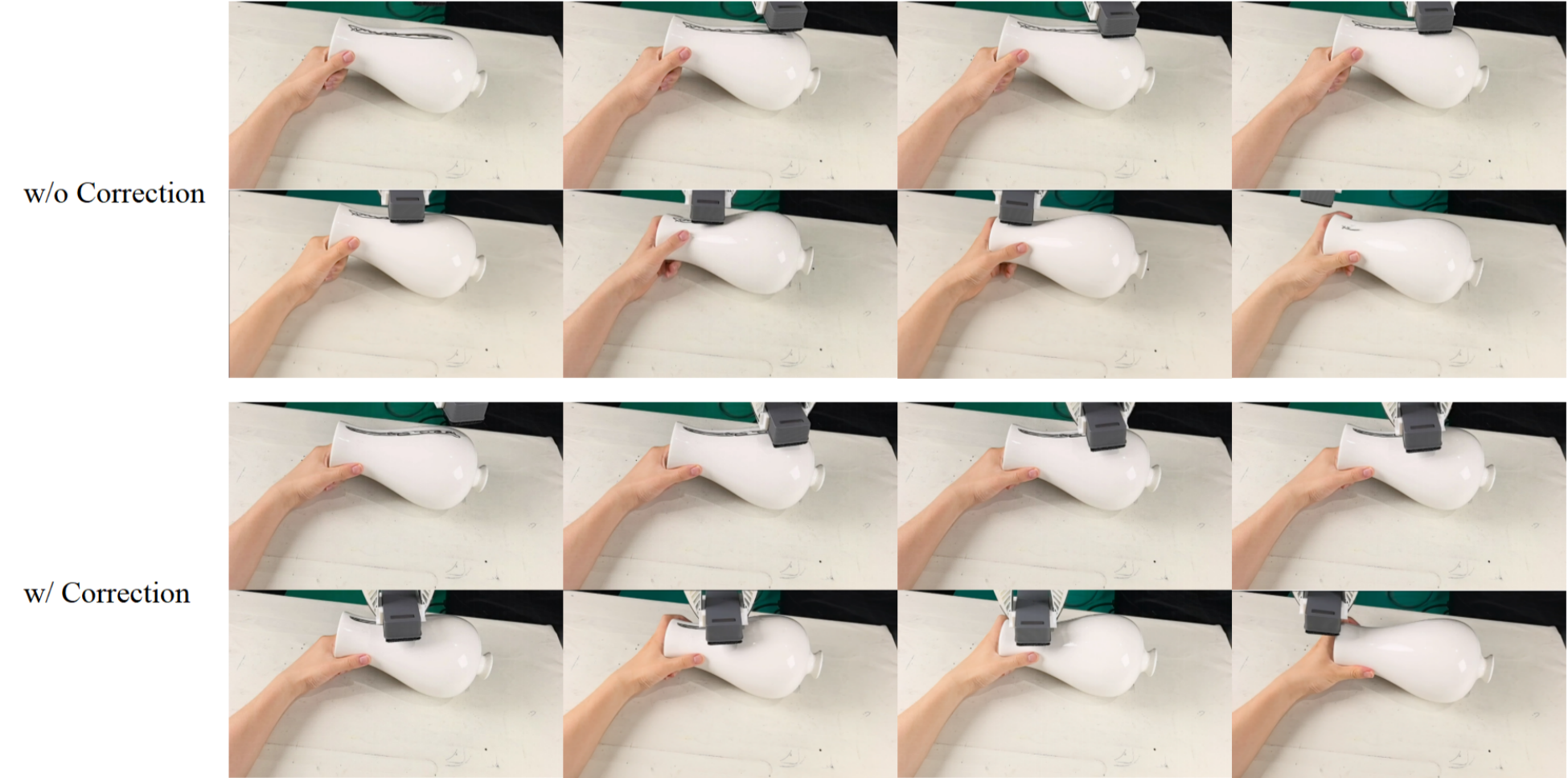}}
    \caption{\textbf{Comparative Correction Experiments for Wipe Vase.} }
    \label{fig:appendix_video_snapshot_4}
\end{figure}

We provide additional qualitative visualizations for the perturbation and correction experiments in Figures~\ref{fig:appendix_video_snapshot_1}--\ref{fig:appendix_video_snapshot_4}.

\newpage
\newpage

\subsection{Additional Visualization Supplement}
\label{app:ablation_qualitative}
\begin{figure}[H]
    \centering
    \includegraphics[width=0.85\linewidth]{\detokenize{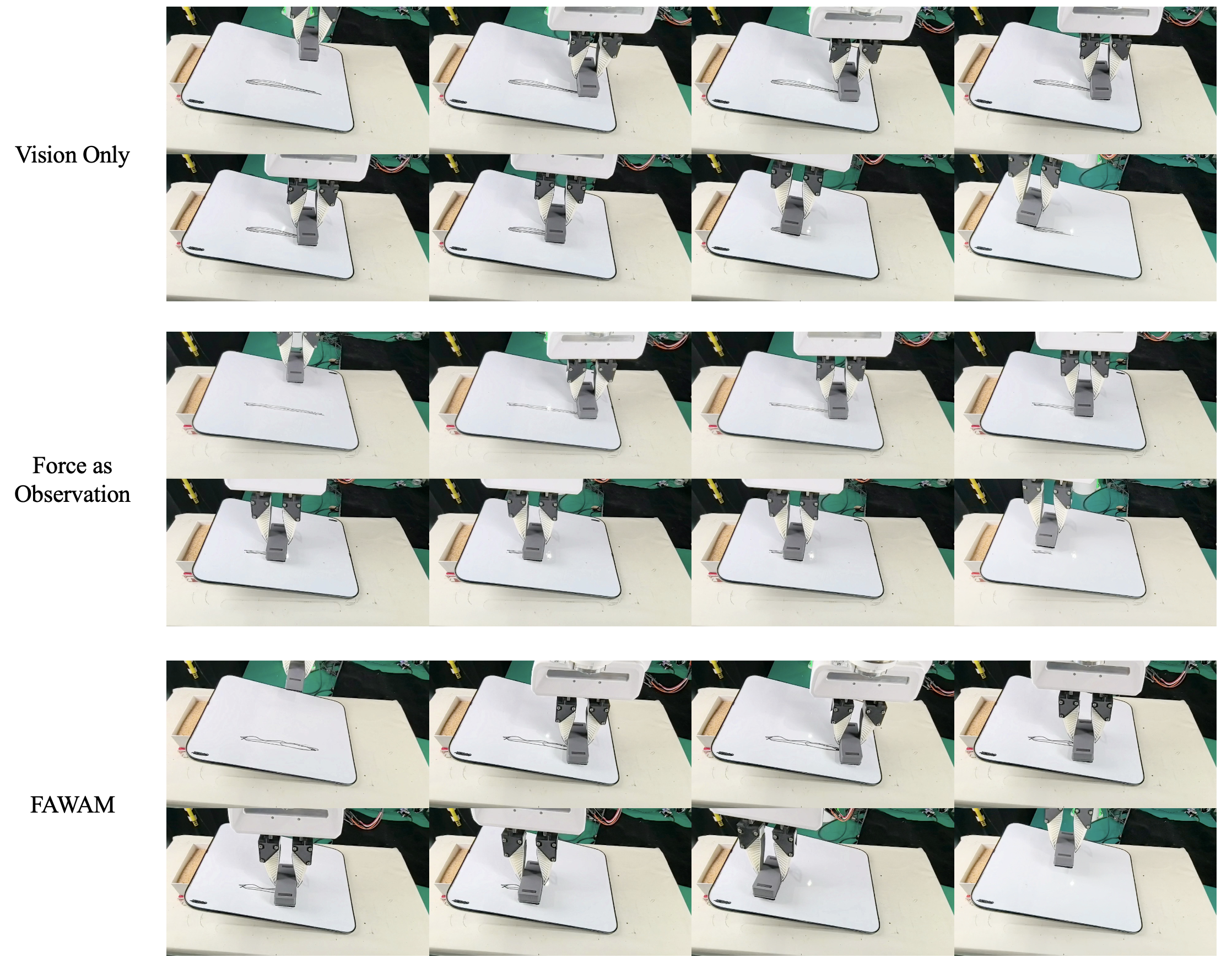}}
    \caption{\textbf{Ablation qualitative supplement for \textit{Erase Board}.} }
    \label{fig:appendix_ablation_wipeboard}
\end{figure}

\begin{figure}[H]
    \centering
    \includegraphics[width=0.85\linewidth]{\detokenize{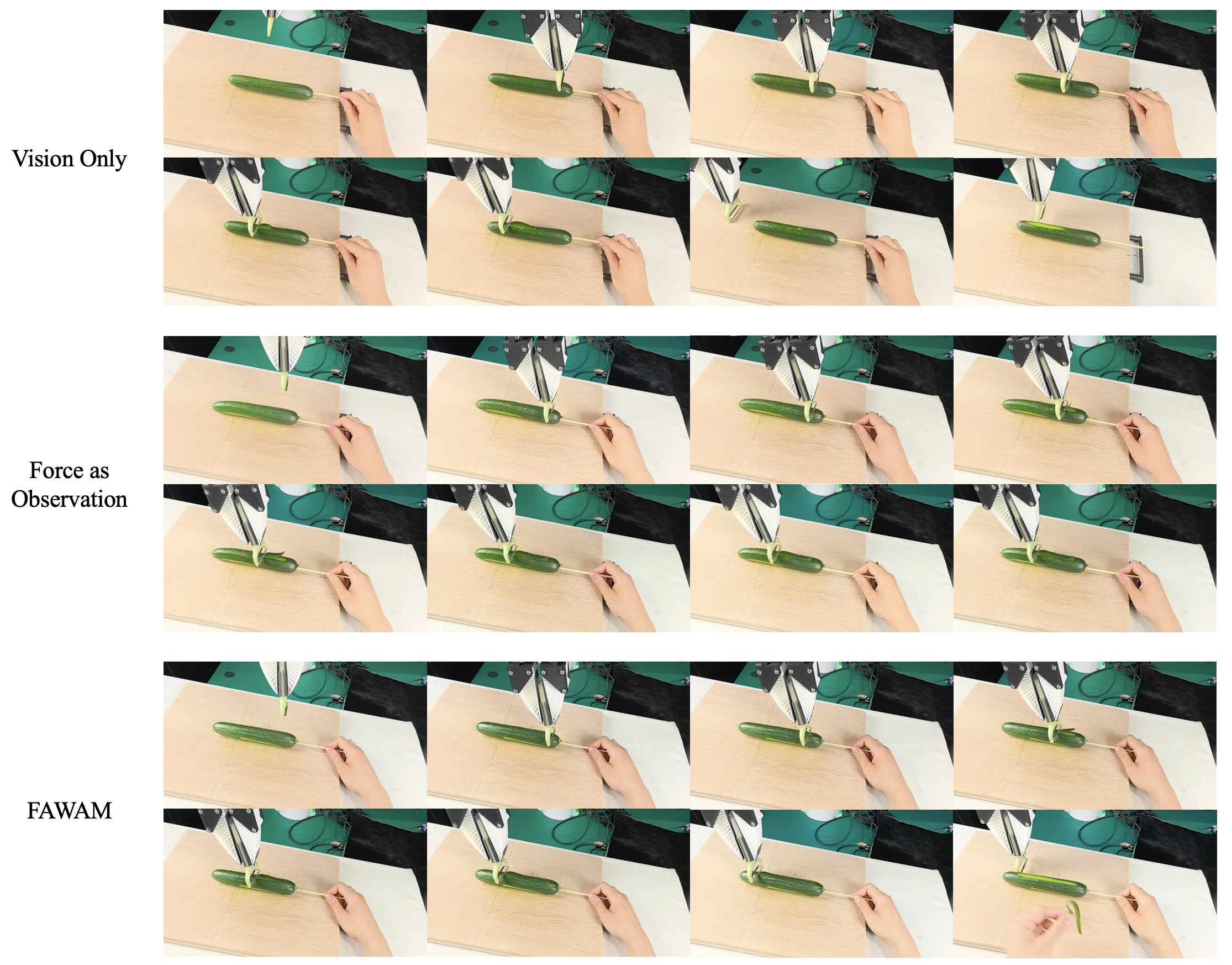}}
    \caption{\textbf{Ablation qualitative supplement for \textit{Peel Cucumber}.} }
    \label{fig:appendix_ablation_peelcucumber}
\end{figure}

\begin{figure}[H]
    \centering
    \includegraphics[width=0.85\linewidth]{\detokenize{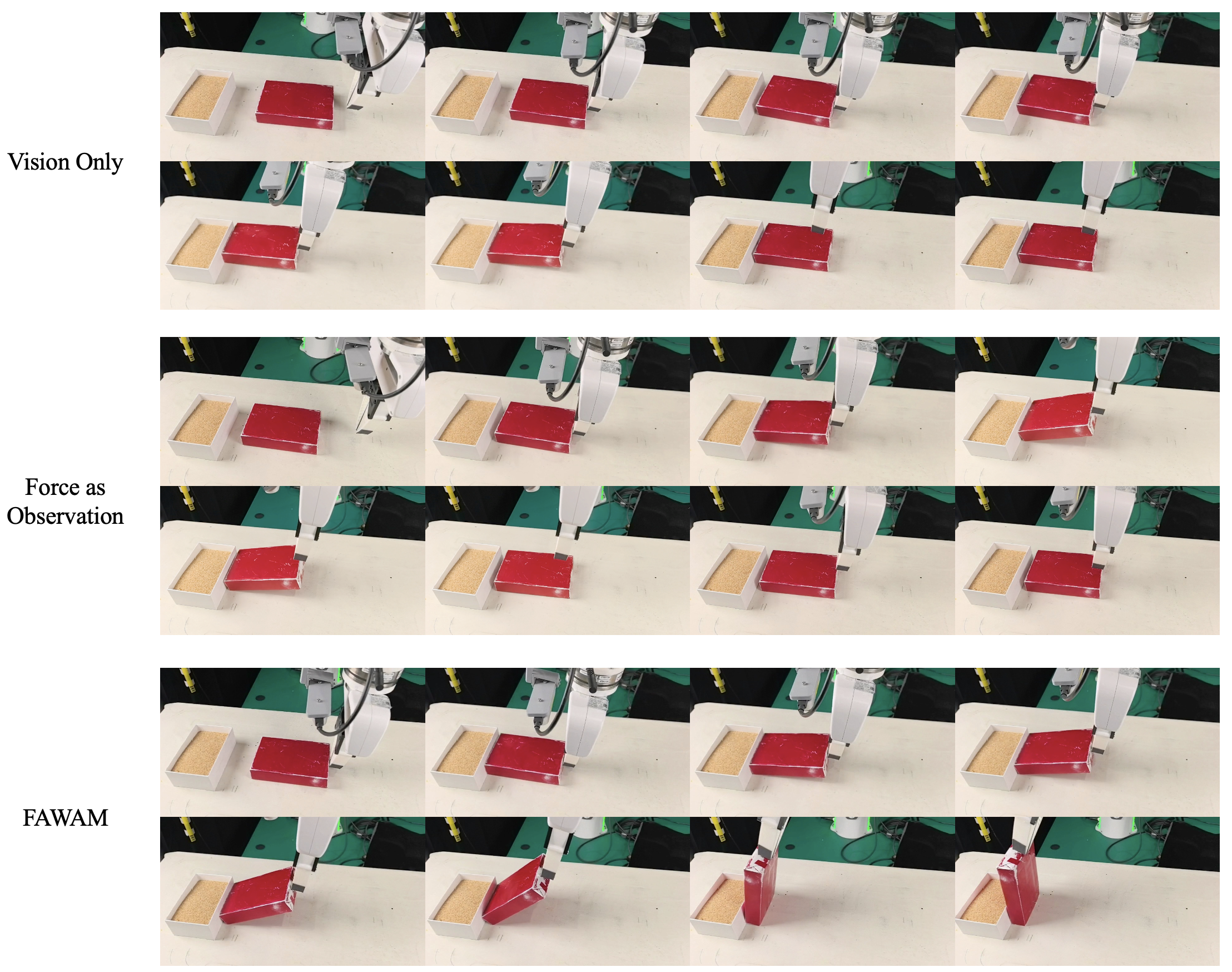}}
    \caption{\textbf{Ablation qualitative supplement for \textit{Pivot Box}.} }
    \label{fig:appendix_ablation_pivotbox}
\end{figure}

\begin{figure}[H]
    \centering
    \includegraphics[width=0.85\linewidth]{\detokenize{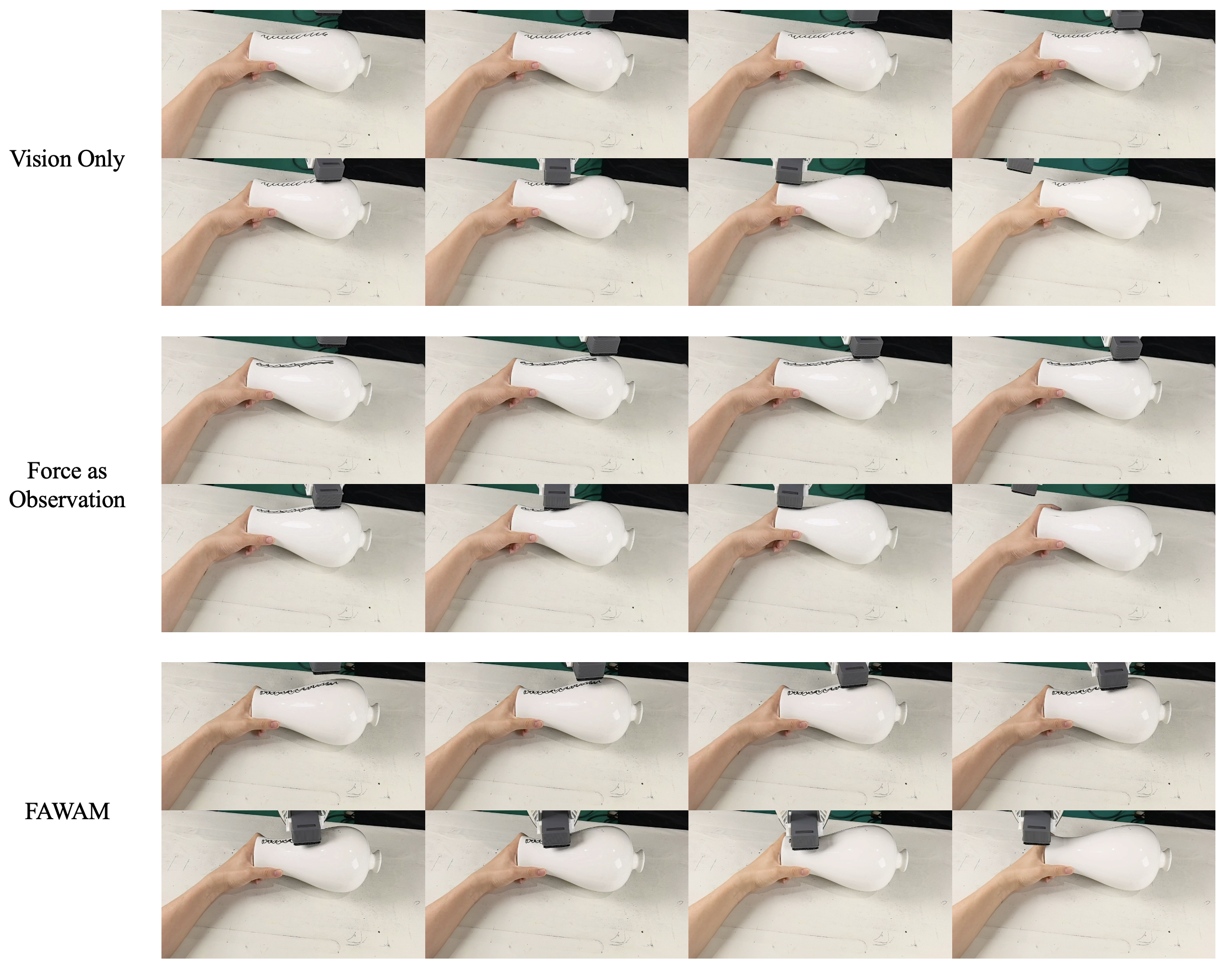}}
    \caption{\textbf{Ablation qualitative supplement for \textit{Wipe Vase}.} }
    \label{fig:appendix_ablation_wipevase}
\end{figure}

Figures~\ref{fig:appendix_ablation_wipeboard}--\ref{fig:appendix_ablation_wipevase} complement the ablation study in Table~\ref{tab:ablation_study} with qualitative visual examples. They present representative executions across the four contact-rich tasks, illustrating the role of joint force-action prediction and online residual correction in maintaining stable contact, adapting to local geometric changes and recovering from interaction errors.
\newpage

\newpage

\newpage
\subsection{Additional Analysis on Residual Activation Threshold}
\label{app:residual_threshold}
\begin{wrapfigure}{r}{0.45\linewidth}
\centering
\includegraphics[width=\linewidth]{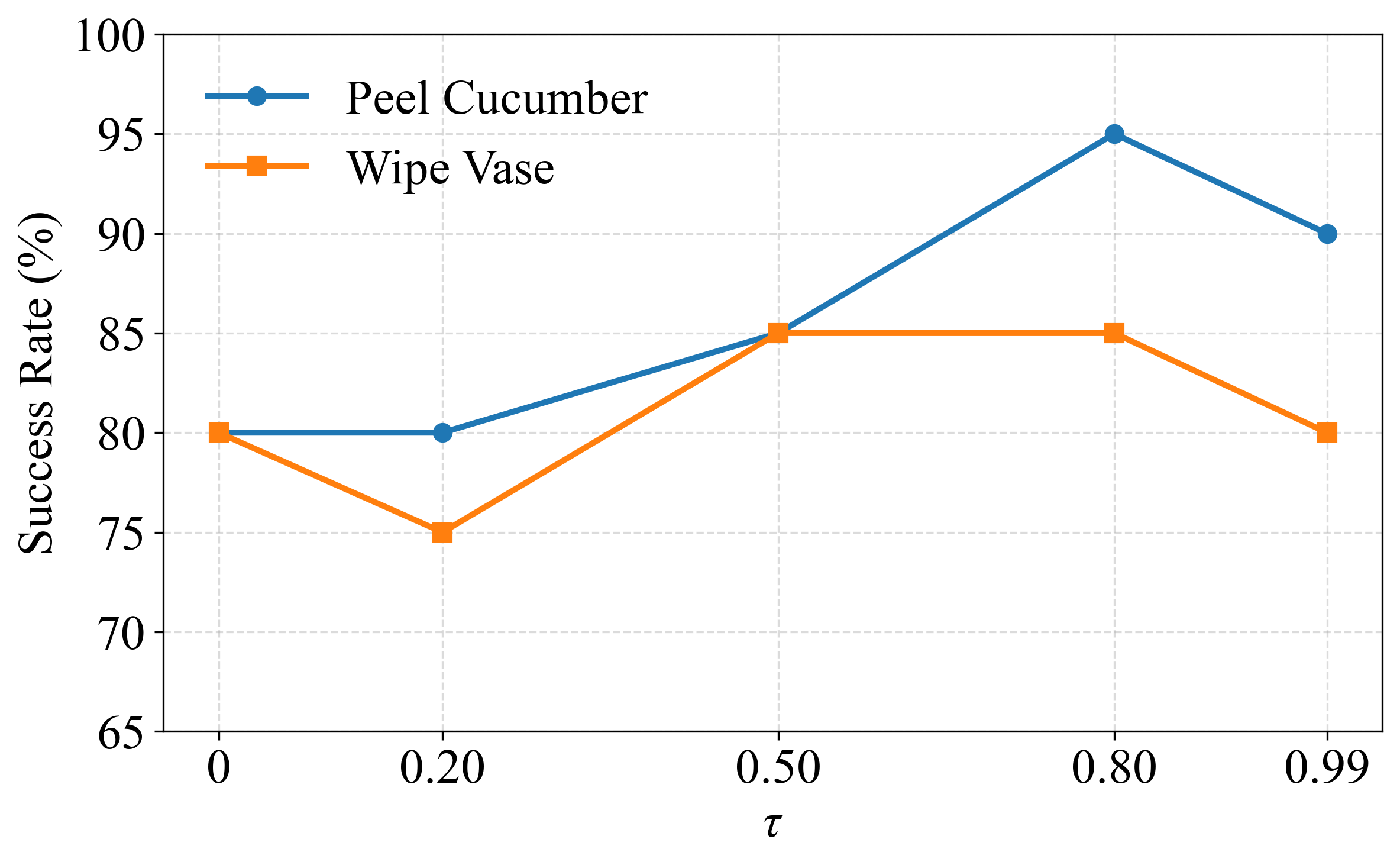}
\caption{Success rate under different $\tau$.}
\label{fig:ablation_tau}
\end{wrapfigure}
In this analysis, we evaluate the effect of different residual activation thresholds $\tau$. A smaller $\tau$ makes the residual gate easier to activate, so residual corrections are more frequently added to the base action chunk. We test $\tau \in \{0, 0.2, 0.5, 0.8, 0.99\}$ on the \textit{Peel Cucumber} and \textit{Wipe Vase} tasks. As shown in the Figure~\ref{fig:ablation_tau}, overly small thresholds lead to lower success rates, suggesting that frequent residual activation may unnecessarily perturb nominal base-policy execution. Increasing $\tau$ improves performance by making correction more selective. In particular, $\tau=0.8$ achieves the best or near-best performance on both tasks, indicating that sparse residual activation is beneficial. However, when $\tau$ is too high, the residual branch becomes overly conservative and may fail to trigger necessary corrections, leading to a performance drop. Overall, these results suggest that using a relatively high but not overly strict activation threshold is important for effectively exploiting the residual correction module.
\end{document}